\documentclass[letterpaper, 10 pt, conference]{ieeeconf}  % Comment this line out if you need a4paper

\IEEEoverridecommandlockouts                              % This command is only needed if 
\usepackage[utf8]{inputenc}
\usepackage{hyperref}
\usepackage{amsmath}
\usepackage{graphicx}
\usepackage{bm}

\usepackage{color}
\usepackage{xcolor,colortbl}

\usepackage{amssymb}
\usepackage{afterpage}
\usepackage{enumerate}
\usepackage{booktabs}
\let\labelindent\relax
\usepackage{enumitem}

\usepackage{adjustbox}
\usepackage{array}

\usepackage{flushend} %to balance columns on last page

 \usepackage{cite}
\newcommand{\Pie}{$\Pi_e$}

\usepackage[normalem]{ulem} %strike through

\title{\LARGE \bf
Encoding cloth manipulations using a graph of states and transitions
}

\author{Júlia Borràs, Guillem Alenyà and Carme Torras% <-this % stops a space
\thanks{The research leading to these results receives funding from the European Research Council (ERC) from the European Union Horizon 2020 Programme under grant agreement no. 741930 (CLOTHILDE: CLOTH manIpulation Learning from DEmonstrations) and  is also supported by the Spanish State Research Agency through the María de Maeztu Seal of Excellence to IRI (MDM-2016-0656) and the Spanish Ministry of Science and Innovation, project HuMoUR (TIN2017-90086-R). }
	\thanks{The authors are  with Institut de Robòtica i Informàtica Industrial, CSIC-UPC,  
		Llorens i Artigas 4-6, 08028 Barcelona, Spain. {\tt \{jborras, galenya, torras\}@iri.upc.edu}}%
}

\begin{document}

\maketitle
\thispagestyle{empty}
\pagestyle{empty}

%%%%%%%%%%%%%%%%%%%%%%%%%%%%%%%%%%%%%%%%%%%%%%%%%%%%%%%%%%%%%%%%%%%%%%%%%%%%%%%%
\begin{abstract}
Cloth manipulation is very relevant for domestic robotic tasks, but it presents many challenges due to the complexity of representing, recognizing and predicting the behaviour of cloth under manipulation. In this work, we propose a generic, compact and simplified representation of the states of cloth manipulation that allows for representing tasks as sequences of states and transitions. We also define a Cloth Manipulation Graph that encodes all the strategies to accomplish a task. Our novel representation is used to encode two different cloth manipulation tasks, learned from an experiment with human subjects with video and motion data. We show how our simplified representation allows to obtain a map of meaningful motion primitives.% and to segment the motion data to obtain sets of trajectories, velocity and acceleration profiles corresponding to each manipulation primitive in the graph.
\end{abstract}

%%%%%%%%%%%%%%%%%%%%%%%%%%%%%%%%%%%%%%%%%%%%%%%%%%%%%%%%%%%%%%%%%%%%%%%%%%%%%%%%
\section{Introduction}

%Cloth manipulation is an important area of robotics research that has applications both in industrial scenarios and in domestic environments.  Despite its importance, research efforts have been focused mostly on manipulation of rigid objects, showing a great progress in service robotics \cite{torras2016service}, while core capabilities such as grasping, placing, or handing to a person still remain as a hard and unsolved problem when dealing with challenging objects such as textiles. Recently, a stronger interest in deformable object manipulation emerged and the survey in \cite{sanchez2018robotic} presents the latest advances.
 Cloth manipulation presents many additional challenges with respect to rigid object
 manipulation. % related to the complexity of modeling the object and predicting its behavior \cite{jimenez2017visual}. 
 %For this reason, cloth manipulation research has traditionally put more effort in cloth state estimation and grasp point detection, whereas manipulation skills are underdeveloped \cite{sanchez2018robotic}. 
%For instance, one of the crucial skills for manipulation is grasping, and there are very few works analyzing grasping for highly-flexible objects, despite having a variety of grasp types is of paramount importance to determine possible actions and define the state of a scene in terms useful for manipulation. In this direction, we presented a taxonomy of textile grasps \cite{borras2019grasping} based on the geometry of the prehension patches conforming the shape of the grasped part of cloth. 
%In particular, understanding and recognizing manipulation either performed by a human or by a robot is a difficult problem even with rigid objects. Therefore,
In particular, the complexity of defining and recognizing scene states dealing with clothes makes getting reliable data very difficult, hindering the training of AI systems and task planners.

\begin{figure}[bt]
    \centering
    \includegraphics[width=\linewidth]{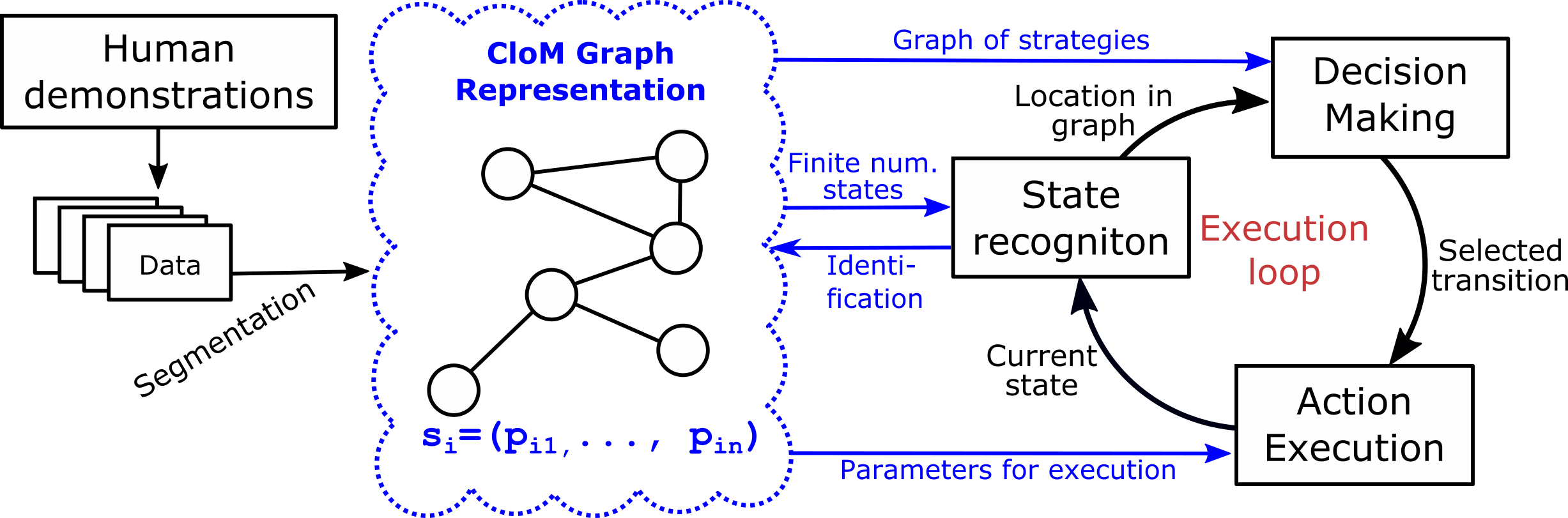}
    \caption{Generic pipeline for learning from human demonstration for manipulation tasks. A good task representation learned from the segmentation of the data can be used for decision making.  State representations have to be defined to ease state recognition but also to enable action execution.
    }
    \label{fig:pipeline}
\end{figure}

Although learning techniques can benefit from simulation,
the transfer to reality has only been successful for simple skills~\cite{matas2018sim,yan2020learning,hoque2020VisuoSpatial,tanaka2018emd}, because simulated cloth differs highly from real behaviour. 
There have been some works learning from real data using either video and sensory-motor data from a robot performing the manipulation in teleoperation~\cite{yang2016repeatable} 
%or reinforcement learning with initial kinesesthatic teaching that evaluating the quality of the final fold result as a reward function \cite{colome2018dimensionality}, 
or from demonstrated robot actions connecting different images of the scene \cite{lippi2020latent}. However, they show clear limitations when it comes to generalizing to other tasks  \cite{yang2016repeatable} or when the scene contains cloth with self-occlusions \cite{lippi2020latent}.  It is even less common to learn cloth manipulation tasks from human demonstrations. However, learning from humans would be important to   obtain  a diversity of strategies to accomplish a task, and with different parameters related to safety, fast accomplishment of the objective or number of steps needed to accomplish a task, inducing a measure of task complexity. 
Learning through human demonstration follows a pipeline similar to \autoref{fig:pipeline}. Large amounts of data could be obtained from human demonstrations in the form of video data and motion data of the hands \cite{verleysen2020video}, but learning from this kind of data is challenging due to the difficulty of annotating data and recognizing cloth states from images. 

Another challenge for cloth manipulation is to find general solutions~\cite{sanchez2018robotic}.
Most recent attempts to find  general approaches consist in end-to-end  learning approaches \cite{tanaka2018emd,yan2020learning, hoque2020VisuoSpatial, lippi2020latent} that are still limited to relatively simple tasks with limited self-occlusions, that is, with fabrics laying flat or semi-flat on a table. We believe the key into general solutions is to define an efficient scene state representation (cloud box in \autoref{fig:pipeline}) to facilitate state recognition but including  sufficient parameters for decision-making and action execution.
%The literature of cloth manipulation has dealt since the very beginning on how to represent cloth and simplify it to a tractable manner to plan actions \cite{miller2012geometric,Doumanoglou2016}. 
%In \autoref{sec:relatedWork} we will review how the literature has deal with the problem of scene representation in the context of cloth manipulation.
% Approaches range from very simple 1-dimensional representations \cite{petrik2016physics} to mesh representations of the cloth [\cite{bersch2011bimanual}], including to  end-to-end learning approaches using the whole scene state image as state definition \cite{yan2020learning}. 

The first contribution of this work is to propose a novel idea to define the scene state in cloth manipulation tasks. The novelty lies in including information on how the  cloth is grasped \cite{borras2020grasping}, where it is grasped from, what are the environmental contacts and the possible transitions between them. 
The second contribution is the Cloth Manipulation (CloM) Graph, a graph that can be built using the previous representation to encode all the possible states and transitions of a given manipulation task seen from video demonstrations, enabling to capture the diversity of strategies. We show the feasibility of our approach extracting the graph for two textile manipulation tasks, one of folding a napkin in 3 folds and another to unfold and put a tablecloth, following a recent benchmark~\cite{garcia2020benchmarking}. We performed an experiment with 8 subjects that are wearing a gripper and the Xsens suit. %(\autoref{fig:setup}). 

From the motion data and the labelling of video data we extract a map of strategies, and we generate segmented motion data that contains specific parameters about the arms trajectories, velocities and accelerations of each segment. % that have been used for each identified manipulation primitive. %
The labelled video data and segmented motion data will be made public, enabling the training and comparison of state recognition algorithms.  %Question: especifiquem que no farem res amb el motion data?

The proposed scene representation and the CloM Graph is also motivated to potentially provide explainability to the decision-making processes, in line with the trustworthy AI from the EU guidelines. As opposed to opaque end-to-end deep learning methods \cite{yang2016repeatable,tanaka2018emd}, latent space variables \cite{lippi2020latent} that are difficult to interpret, or learned latent dynamic models from large amounts of random samples \cite{yan2020learning} that produce plans that are difficult to explain to a human, our CloM Graph provides a framework that is designed to provide both semantic explanations by construction as well as low-level building blocks to plan a task and execute it.

\section{Related work} \label{sec:relatedWork}

Task planning understood as a decision-making module that evaluates different strategies and chooses the optimal plan has been quite unexplored in cloth manipulation. %  due to the difficulties of representing scene states and tasks as a whole.
Seminal literature on cloth manipulation was more focused on motion planning given a task plan \cite{cusumano2011bringing,Doumanoglou2016}. For cloths already flat on a table, simplified planar polygonal representations were used in \cite{miller2011parametrized,Doumanoglou2016,li2015folding} or even simpler 1-dimensional ones in~\cite{petrik2016physics} for rectangular clothes. For grasping hanging clothes, contours were used in \cite{triantafyllou2016geometric}.

Recent more general literature has focused on deep learning approaches where the scene is represented as RGB-D images and the system learns the mapping between an image with an action and a resulting image, where the action is modelled as the pick-up point pixel coordinates and a direction of displacement \cite{hoque2020VisuoSpatial,seita2018deep,yan2020learning,jangir2020dynamic}. In \cite{matas2018sim} they apply reinforcement learning, where the state is represented by an RGB image plus the robot arm joints and grippers state. All these works are trained in simulation but achieve acceptable sim-to-real results. % In\cite{jangir2020dynamic} can even learn dynamic motions. 
In \cite{yang2016repeatable} they use a similar approach by feeding directly the RGB image and robot arm joints to a neuronal network that is trained with teleoperated real robot data.

A few works do task planning using similar approaches. In \cite{tanaka2018emd} they use deep learning to obtain mappings between image states and sequences of simple actions. The method is general but only achieves very simple plans due to the large amounts of data needed, that are in simulation but augmented with large amounts of real robot data.  The work in~\cite{lippi2020latent} is, up to our knowledge, the only that considered the importance of building a graph of scene states to enable task planning. They build a graph in latent space where each node is a set of RGB images related by just perturbations that is linked to another node if it can be obtained through the application of a simple action, modelled as pick-up point and release point in pixel coordinates. The system is trained by demonstrating the linking actions with a real robot.

All these works assume the basic scene state is the cloth when is not touched by the robot. Instead, in our approach, every re-grasp, contact with the environment or change in cloth configuration triggers a new segment in the graph. We believe this is necessary to approach complex tasks where several re-grasps are needed before the cloth is fully released, to obtain simpler action primitives that can be reused in different tasks and contexts, similarly as it was done for rigid objects \cite{zoliner2005towards}. To the best of our knowledge, no work has been able to learn from videos of human demonstrations.

In \cite{jia2019cloth} they do imitation learning in robot-human collaboration tasks. They assume the scene is the RGB-D image and the N coordinates of the points where the cloth is grasped, and define the action as the destination location of the grasped points. In this case, no re-grasp or release is considered.

High-level planning has been tackled in the context of robot-assisted dressing \cite{canal2018joining, kapusta2019personalized}, but without addressing the cloth representation issue and minimizing the part of cloth manipulation by assuming pre-grasped garments.

In our previous work \cite{borras2020grasping} we introduced a framework to describe textile grasps based on the geometry of the prehension agents, including extrinsic geometries from the environment. In this paper, we use that notation to identify the grasp but we use additional information to define the scene state. %afegit pq ho demanava un reviewer

\begin{figure}[tb]
    \centering
    \includegraphics[width=\linewidth]{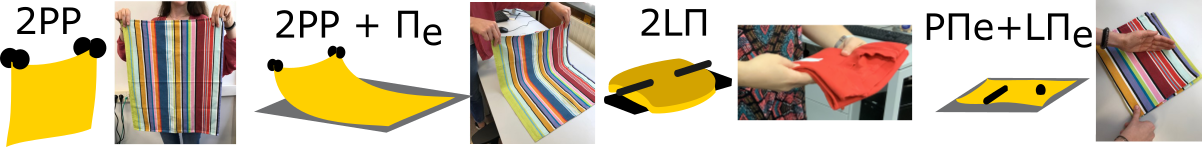}
    (a) \hspace{0.20\linewidth}(b) \hspace{0.20\linewidth}(c) \hspace{0.20\linewidth} (d)
    \caption{The geometries of prehension are points (P), lines (L) and planes ($\Pi$). (a) Double pinch grasp.  (b) A double pinch with the additional extrinsic contact of the table, denoted with an "e" subscript. (c) A double line-plane grasp (d) A combination of grasps of the hands against the table.}
    \label{fig:graspFramework}
\end{figure}

\begin{figure}[bt]
    \centering
    \includegraphics[width=0.6\linewidth]{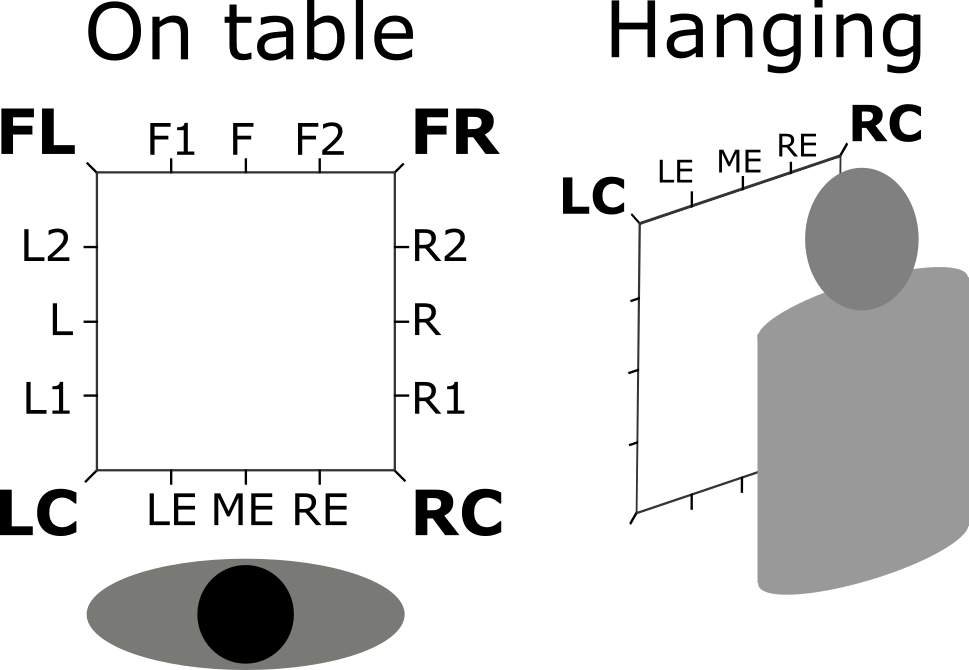}
    \caption{Location of grasp points with respect to subject. Any interior point is labeled I.}
    \label{fig:graspLocations}
\end{figure}

\section{A generic state-and-transition definition} \label{sec:definitionOfStateAndTransition}

To recognize and understand a manipulation action, it is necessary to interpret the states of a scene at each time-step. This is a difficult problem and our approach is to define a simplified representation of a scene in a way that can be recognized by a robot and that allows executing the next action.

% \begin{figure*}[tb]
%     \centering
%     \includegraphics[width=\linewidth]{figs/compilationFrames.png} 
%     \caption{Above, snapshots of the video data labeled according to our four-field characterization of manipulation primitives, all at the beginning of a transition. Below, the corresponding scene representation automatically generated from the labels.}
%     \label{fig:labeledScreenshots}
% \end{figure*}

We propose to define a  {\it state} as a tuple  $$\textbf{S}=< GT, GL, CC>$$ where 
 \begin{itemize}
     \item $GT$ is the grasp type,
     \item $GL$ are the location of the grasp with respect to the cloth, and
     \item $CC$ is the cloth configuration.
 \end{itemize}

Then, we define a  {\it manipulation primitive} as the triple $$< \textbf{S}_o, \textbf{S}_d, M>$$ where  
 \begin{itemize}
     \item $\textbf{S}_o$ and  $\textbf{S}_d$ are the origin and destination states, and
     \item $M$  is a semantic label of the action primitive the subject is performing.
 \end{itemize}

The definition of the grasp type $GT$ is based on the cloth grasp framework and taxonomy introduced in our previous work \cite{borras2020grasping}. In this framework, each grasp is defined by the geometries of the two virtual fingers that apply opposing forces. %, that is, the geometries of each virtual finger \cite{iberall1986opposition, iberall1991parameterizing}. 
%The framework considers as virtual fingers geometries that can be either intrinsic (part of the gripper) or extrinsic (like the table). %It also considers bi-manual grasps in a natural way. 
A partial glimpse of the grasp framework is provided in \autoref{fig:graspFramework}. A very important feature is that our grasp framework considers elements in the environment as extrinsic contact geometries and, therefore, it explicitly models environmental contact interactions. Thus, all cloth states realize a grasp, as when there is no contact with the subject, the cloth lays on a table, corresponding to a non-prehensile \Pie{} grasp. %, or it is held by other elements in the environment, such as for example a hook, i.e., a \Pe{} grasp, or a rail, i.e., a \Le{} grasp, all of which relying on gravity as the opposing force.

\begin{table}[t]
    \caption{Example frames by scene state }     \label{tab:statesTable}
       \includegraphics[width=\linewidth]{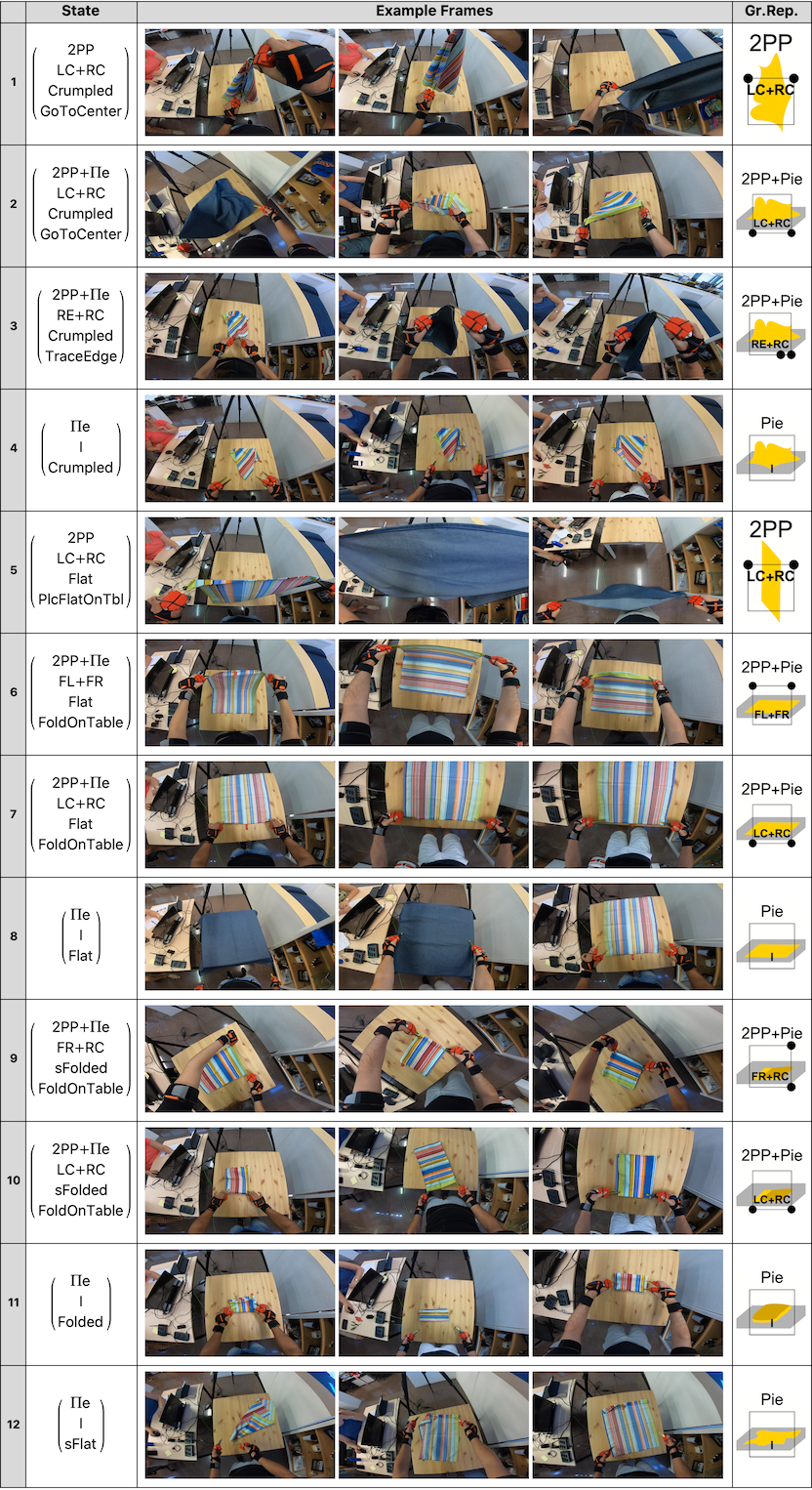} 
       All frames, including frames from additional states, can be found in the paper website
       http://www.iri.upc.edu/groups/perception/\#CloMGraphPaper
\end{table}

%Given a manipulation, information on the grasp at each time-step already provides a segmentation of the action in terms of re-grasps. However, grasp type information alone is not enough to describe the state of a scene. 
%First, relevant information on the cloth state may be missing. For instance, the grasp in \autoref{fig:graspFramework}(c) is with a crumpled cloth, but it could also happen with a flat cloth on the table. Second, the location of the grasping point on the cloth is also relevant, because one can have a \PP{} grasp on a corner, an edge or an interior point, which imply very different action possibilities. Finally, for an understanding of the scene, it is also important to have a semantic interpretation of the action being performed stemming from the combination of grasp type, grasping point locations and cloth state.
Regarding the grasp location $GL$,
 we have defined a set of labels to describe the approximate locations of the grasping points on a given rectangular cloth, shown in \autoref{fig:graspLocations}, corresponding to coordinates in a 2D cloth reference. Note that a similar notation could be used for other shaped garments. Locations are encoded with respect to the subject grasping hands, i.e., left corner (LC) refers to the corner closest to the subject at that side, and right corner (RC) likewise, up to rotations of 45ª. The two farthest corners are labelled far left (FL) and far right (FR). When the cloth is hanging, the right and left corners are the top ones (closer to subject hands). This means that for certain state transitions we may get a swap of labels for the same points. For instance, when placing a cloth flat on a table, and then folding it without releasing it, the labelling swaps from (LC+RC) to (FL+FR) after the table contact has been added. See the next section for more details and examples. This notation is used regardless of the cloth configuration. Therefore, when the cloth is folded, each corner contains several layers of fabric. If only the top layer is grasped, it is noted with the subscript RC$_1$. If no subscript is used, it is assumed the subject is grasping all the layers. Although this label represents a coarse grid on the border of the cloth, the associated manipulations that a robot may do depending on these locations does not require more precision, as only the concept is important for the decision process. For the execution phase, additional information could be added regarding the location of the robot grippers.

Regarding the configuration of the cloth, $CC$, it is well known the configuration state of a textile is infinite-dimensional. %, therefore this parameter could be of very high complexity. 
That, together with the high number of self-occlusions that occur when manipulating clothes, makes cloth state estimation a difficult problem. The high complexity of its full solution has been bypassed in the past by just looking for task-oriented features, such as adequate and accessible grasping points, e.g., shirt collars for hanging \cite{ramisa20163d} or towel corners for folding \cite{maitin2010cloth}. %, edge of cap visor to help dress  
 Increasingly, it becomes clearer that we need simplified representations, specially regarding deformable objects, as stated in \cite{smith2012dual}.
%In the framework of the European ERC project Clothilde, we want to simplifthe configuration space in a manipulation-meaningful way, i.e., distinguishing only between cloth states that should be manipulated differently, without having to estimate the complete state of the textile object. \GA{AIXÒ ENTRA MOLT BRUSC}
We have defined only 5 categories of simplified cloth configurations: 
$$\{ Crumpled, Flat, Folded, Semi{\text -}Folded, Semi{\text -}Flat\}.$$
%For the crumpled category, there are subcategories dependent on the number of visible corners. %A further simplification in the current work has been to assume there is always a visible corner that can be grasped.
%This is enough to extract coherent sequences of states for a task. 
This is a very short list of states, but in combination with the grasping information and the interaction with the environment, we found it reduced the variability enough inside one same state. This can be seen in \autoref{tab:statesTable}, where we show examples of frames corresponding to segments identified in our experiments. For instance, the crumpled state, that appears in rows 1-4, can have many configurations. However,  whether it is in contact with the table or not, or grasped by corners or not reduces the possible configurations to very similar shapes inside each state. This is not true for the case where it is not grasped, like in row 4. In this case there is the possibility of enriching each category with different descriptors like \cite{ramisa2013finddd} to measure the amount of deformation or the number of visible edges and corners using methods like \cite{qian2020cloth} or \cite{liu2016fashion}. This is out of the scope of this work. 

The other state that may seem ambiguous is the semi-folded, rows 9 and 10 in \autoref{tab:statesTable}, as we are not considering how many folds have been done. Indeed, we can see in the table how cloths with different number of folds appear under the same state. However, we propose to only identify the final state of fold (row 11), as all partial folds afford the same kind of action, that of continue folding until you are done.

The semi-flat state (row 12) is important as it can trigger a flattening action, but it has been purposely ignored in the data for simplicity, as will be explained in the next section.

Finally, regarding the motion semantic label $M$, we define a set of labels related to the action the subject is performing from that initial state until the following one, like for instance, "Place flat on table", "Fold on table" or "Trace edge". Semantic labels are useful for high-level planning and scene understanding, and can be linked to low-level parameters like motion primitives or other trajectory representations. They can also be seen in \autoref{tab:statesTable} because for the data we have collected, all states where the cloth is grasped trigger the same action although they may finish in different states, as can be seen in the graph representations.

The proposed state and transition definition induces a segmentation of manipulation tasks at each change of scene state. The state changes at each re-grasp, which in our grasp framework, this includes changes in contacts with the environment. In addition, there is also a change of state when the grasp locations vary (like in a "Trace Edge") or when the cloth configurations changes (like in "Unfold In The Air").

%For instance, unfolding in the air starts from a 2\PP{} grasp at the corners, with the cloth crumpled, and the two hands move to make the cloth flat (unfolded) by the end of the primitive. If then the subject places the grasped points on the table, this constitutes a different manipulation primitive although the grasp type and grasping point locations haven't changed. For the edge tracing manipulation, the subject performs a pinch-and-slide motion that results in different grasping point locations at the beginning and end of the segment. 

\begin{figure}[tb]
    \centering
    \includegraphics[width=0.6\linewidth]{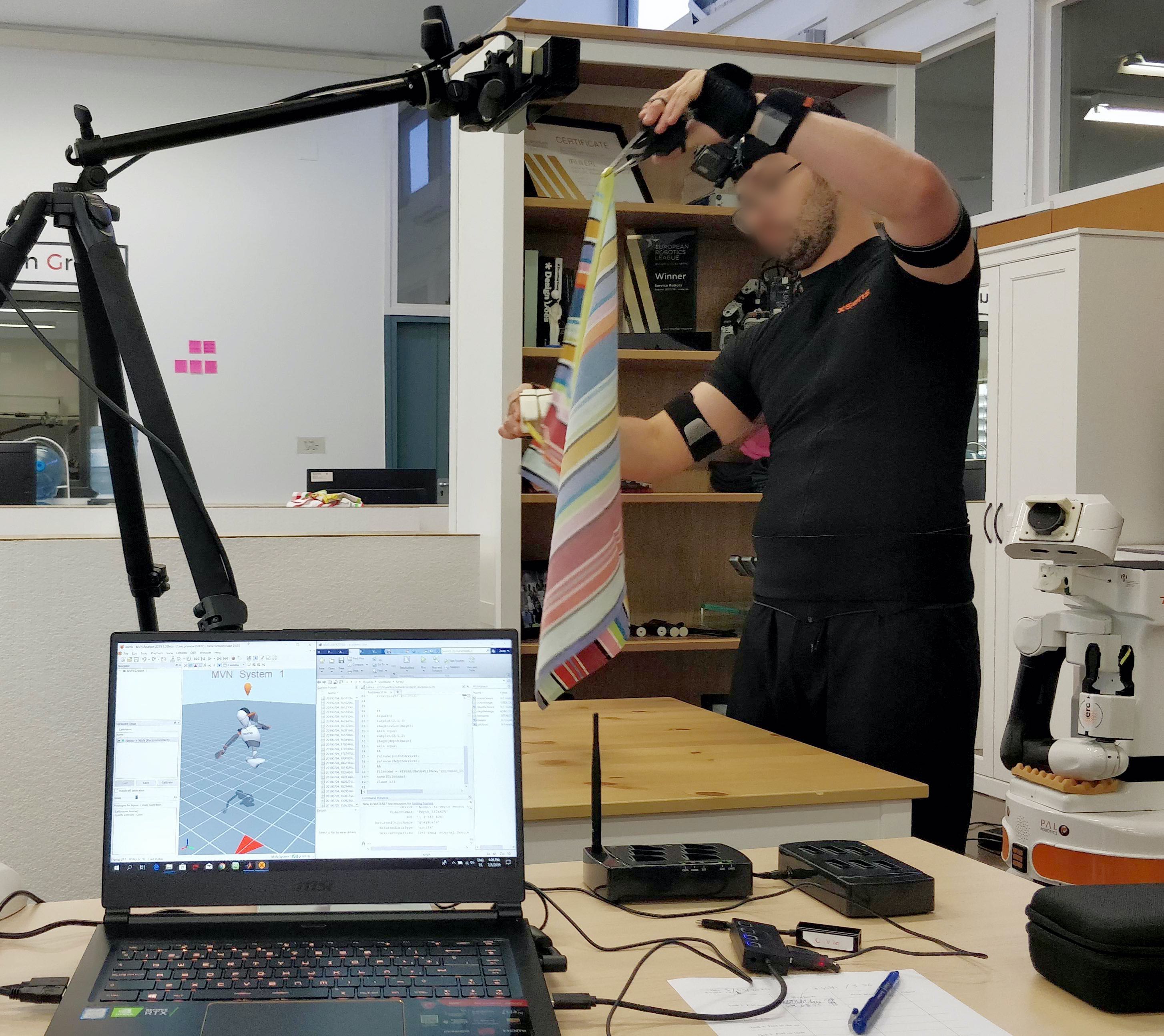}
    \vspace{0.3cm}
    \includegraphics[width=0.4\linewidth]{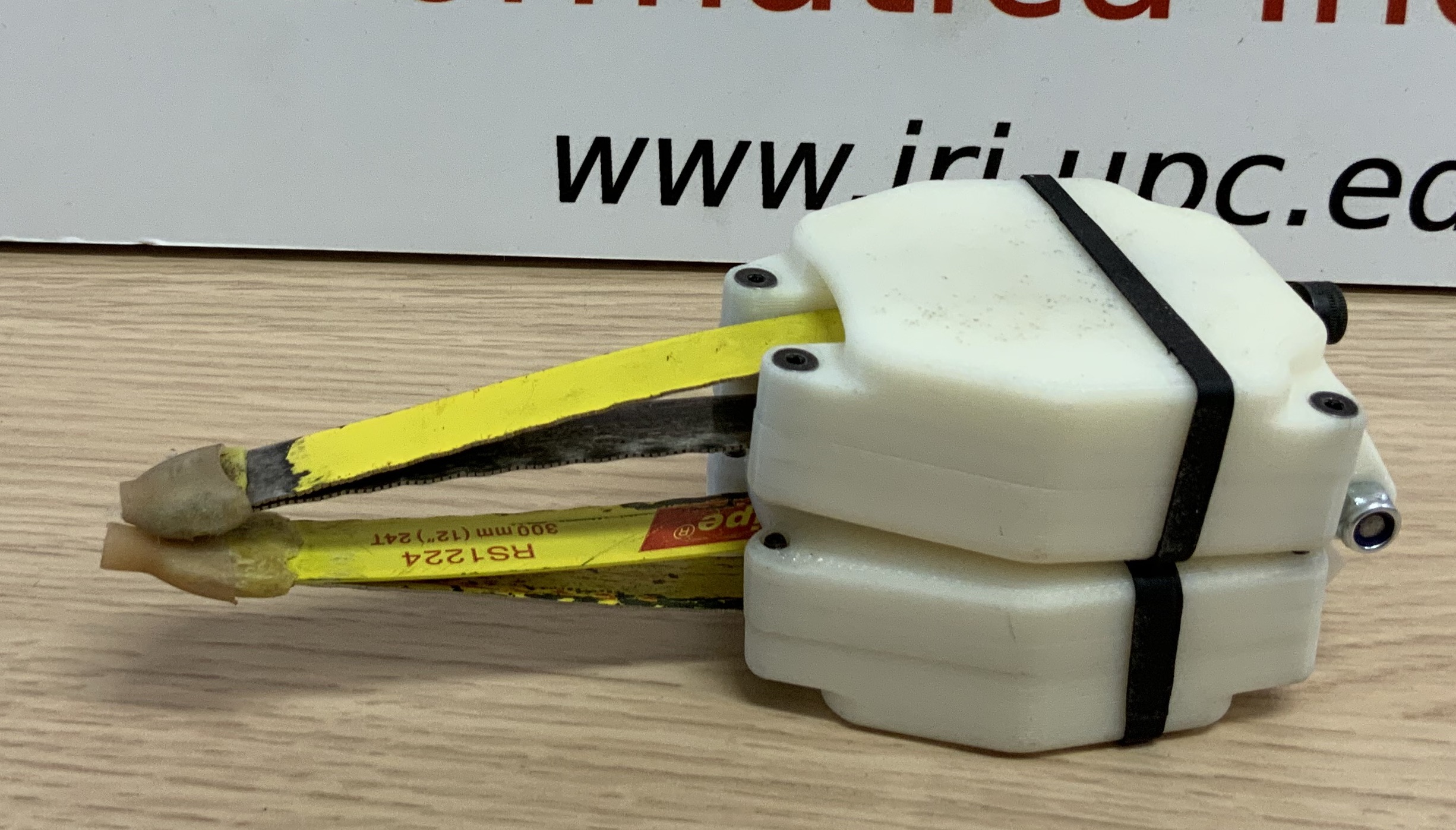}
    \caption{Experimental setup. (Top) The subject wears a motion tracking suit, a GoPro camera mounted on the head and we also take a Kinect screenshot of the final result, although the latter is not used in this paper. (Bottom) Wearable point-point gripper used in the experiments.}
    \label{fig:setup}
\end{figure}

\section{Experimental setup and data collection}
We tested a total of 8 subjects wearing a motion data suit (XSens) and a GoPro camera fixed at their forehead (\autoref{fig:setup}). 
The experiment included several cloth manipulation tasks, but for the scope of this paper, we focus on the task of folding a napkin with 3 folds on the table and the unfolding to put a tablecloth. We asked the subjects to wear a simple gripper, shown at the bottom of \autoref{fig:setup}, to reduce their manipulation dexterity to one closer to that of the robot. Subjects were allowed to train with the grippers, executing the tasks three to four times before starting the recordings.
 
When it comes to cloth manipulation, human experiments provide us with a lot of useful information regarding the variety of strategies to accomplish a task, that is not observed in robot cloth manipulation demonstrations, as analyzed in~\cite{borras2020grasping}. Therefore, learning state sequences from humans will provide us with a much richer graph regarding alternative strategies, and we will be able to learn new manipulation approaches for robots. However, there is a trade-off between obtaining a great diversity of strategies and sparsity on the obtained data derived from particular ways subjects perform one same task.  This is specially true when it comes to cloth manipulation that  almost every subject has its own tricks to fold their clothes. For this reason, we instructed the subjects to perform a very specific task (fold on the table, not in the air, and in 3 folds, and unfold the tablecloth to directly place it on the table). Despite these indications, we obtained a lot of variability, sometimes even between the trials of one same subject. However, some strategies have been used consistently by most of the subjects.

\begin{figure*}[bt]
    \centering
    \includegraphics[width=\linewidth]{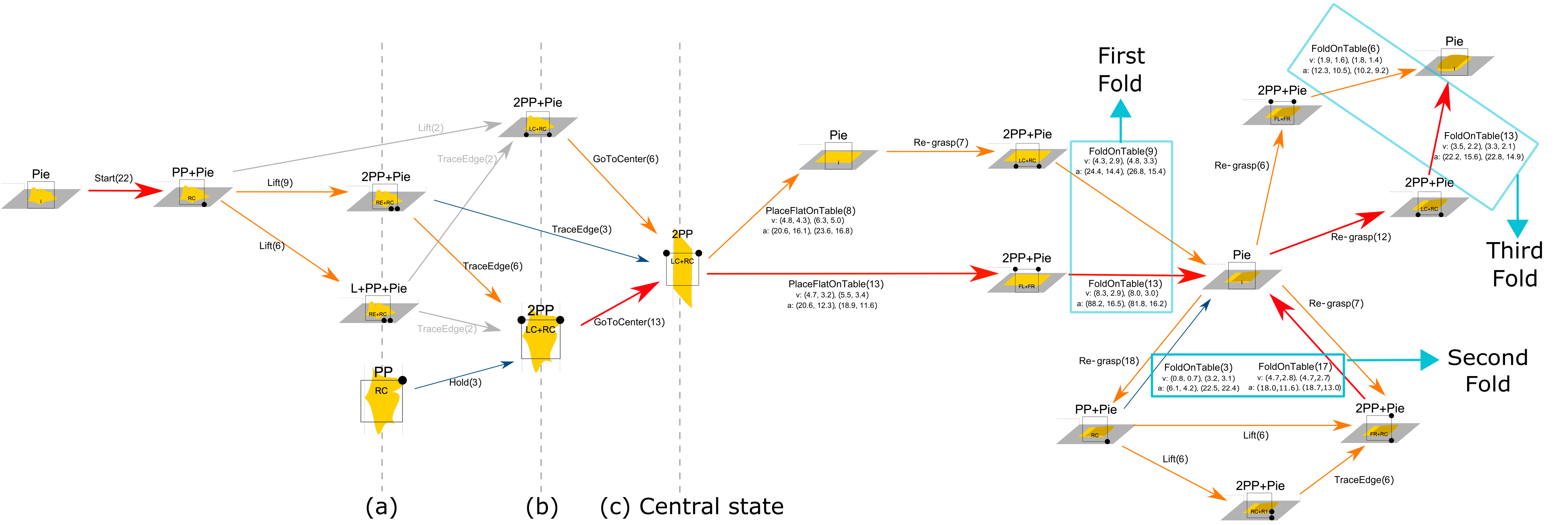}%GraphRestricted_Retocat.pdf} 
    \caption{Reduced CloM graph obtained by requiring each edge to be observed at least 3 times in the data. We can clearly see the different phases of the task, from the crumpled on the table phase on the left, to the central hanging part of the manipulation, and then the semi-folded states on the table during the first, second and third folds, located to the right. The label of each edge consist of the semantic name of the primitive, the number of times it appears in the data (in parenthesis) and for some of them we show the maximum and mean velocities and accelerations averaged for all the trials, for the left and right hand respectively. }
    \label{fig:graphRestricted}
\end{figure*}

From the data collected, we have manually labelled the videos at each change of state, associating a motion semantic label to each transition depending on the action that was done, following the proposed representation. We purposely ignored any manipulation that corrected a mistake, or that relocated the cloth on the table, just to simplify the data. Examples of the labels and their corresponding graphic state representations can be seen in \autoref{tab:statesTable}. 
%We used a subtitle software to generate the labels in text files. 
The labels include timestamps at each change of state, providing the segmentation of the data and the sequence of states. The motion data is synchronized through an initial clapping of the subject, that is labelled in the subtitle and detected as a peak of acceleration in the motion data. In future work, we plan to use the manually annotated data to train a system to autonomously label new video data.

%By using the grasp type framework described before, the grasping point locations in \autoref{fig:graspLocations}, and the 5 cloth configuration labels for the folding task, we generated the scene states by combining these values. 

\section{Cloth manipulations graph}
Thanks to the proposed representation, and extracting the sequences of state and transitions of the labelled video data, we can generate a graph where each node is a scene state, and the edges represent the transition action. %, from origin to destination states, together with the semantic motion label. 
%Thus, the graph represents a map of possible manipulations from the first to the last state. 

To generate the graph, for each trial we defined an edge for each state change, and we represented it symbolically using the formulation introduced in \autoref{sec:definitionOfStateAndTransition}, where each initial and destination states are the initial and end node of the graph edge, and the motion semantic value is the edge label. We then identify common nodes and common edges, defining the graph with all the distinct vertices and edges that have appeared, counting their multiplicity.

We will show two applications of our proposed graph. First, we will show the CloM graph from the task of folding a napkin on the table. Second, we will analyze all the data from both tasks to study a particular transition, that of tracing an edge with a pinch and slide motion.

For the first case, to simplify the data, we have removed some left and right distinctions. For instance, a single corner grasped is the same irrespective of whether it is the left or right corner, grasped with the left or right hand. We also assume two grasped points on the same cloth edge are the same regardless if they are on the right or left side. All these simplifications are described in the additional material \footnote{http://www.iri.upc.edu/groups/perception/\#CloMGraphPaper}.

%As stated before, one of the issues we face with this type of data is that there is a lot of variability in the way individuals perform cloth manipulation tasks, resulting in sparse data. 
Using all the data collected, we obtain a graph with 32 nodes and 65 edges, but many of them appear a single time in our data. If we require each edge to appear at least 3 times in the data, the graph is reduced to 18 nodes with 24 edges. The reduced graph is shown in \autoref{fig:graphRestricted}. The complete graph can't be included in the paper for space reasons, but you can find it on the provided website.  The CloM Graph of the task of unfolding and putting the tablecloth can also be found in the website, in this case, the simplified one has 12 states and 15 transitions, while the full graph has 17 states and 32 transitions, meaning this task is of much less complex than the previous one. As the two tasks are inverse one of the other, only one transition is common in both graphs, the one of "Place flat on table" from the central state (2PP, RL+LC, Flat) to the ($\Pi_e$, - ,Flat) that appears 21 times for the tablecloth task and 8 for the folding task. 

\begin{figure*}
    \centering
    \includegraphics[width=\textwidth]{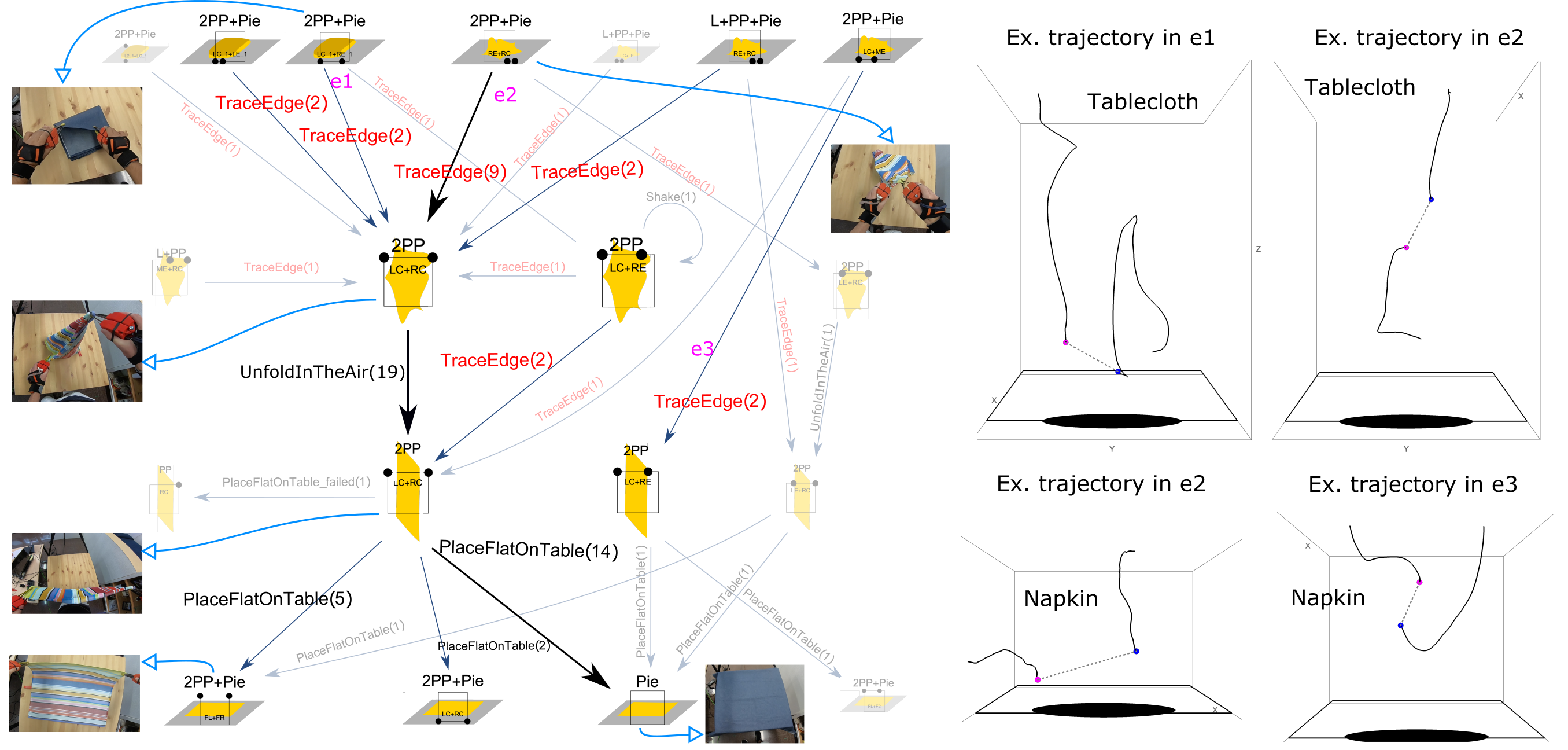}
    \caption{Left: CloM graph of the transition of tracing an edge. Right: Example trajectories of the gripper tip for the transition actions in the edges e1 to e3. Pink dots represent starting points of the left hands and blue of the right hand. Location of the subject and table are approximated.
    %The cyan and blue ellipsoids correspond to left and right hand starting points distribution. The end points of the trajectories from the left and right hands of each trial are linked with a dashed line, that is orange for the executions of task 1 (small napkin) or purple for task 4 (big tablecloth).
    }
    \label{fig:actionGraph2}
\end{figure*}

We performed a total of 24 trials, meaning the maximum times one primitive can appear repeated in the data is 24. Despite the diversity of strategies displayed by the subjects there are some transitions that consistently appear. We plotted in red the transitions that appear in at least half of the total capacity (12 times) and, in orange, the ones that appear 6 times or more.  We can see that the weakest flow in the graph is in the transition from \autoref{fig:graphRestricted}-a to \autoref{fig:graphRestricted}-b. That is because there is a great variety of manipulations to find the two corners, that can be appreciated in the full graph. Once the corners are grasped, the primitives to unfold in the air become less sparse (\autoref{fig:graphRestricted}-c). The bottom state at the column (a), the (PP, RC, Crumpled) state, is reached by several edges with a multiplicity 1 that don't appear, but can be seen in the full graph. 

For the second application, we analysed the skill of tracing an edge. From our data we found two main families of application: The first one, when grasping a single layer of cloth from a flat or folded cloth, the easier corner is first grasped and lightly lifted to then trace the edge towards the other corner. The second family is to achieve the "unfold in the air" and this is the one that we will study in further detail here.

In \autoref{fig:actionGraph2} we show the CloM graph. It is a subgraph of the CloM graph of both tasks, selecting all the edges with label "Trace edge", and their following states until the mid-term objective of placing the cloth flat on the table is reached. In this case, we have not used any of the simplifications from before, except the left and right identification. Those transitions that are sparse in our data appear faded out. On the right of \autoref{fig:actionGraph2} we can see example trajectories of the gripper tips executing the transitions of some of the edges. The complete set of trajectories can be seen in the additional material\footnote{http://www.iri.upc.edu/groups/perception/\#CloMGraphPaper}. %For better visualisation, we translated the trajectories to the same starting point corresponding to the mean of all the starting points, and the ellipsoids show the distribution of the starting points.

We have chosen to analyze this transition because it is very common in human cloth manipulation demonstrations, but it hasn't been very studied in the context of robotics. Only one paper uses it applied to very small clothes \cite{yuba2017unfolding}, but several grippers have been designed to ease this task \cite{sahari2010edge, shibata2012fabric, donaire2020versatile}. 

We can extract several conclusions from the Trace Edge CloM graph. First, we can observe that for most of the cases, during edge tracing the subject also removes contact with the table. This can be seen in the trajectories of e1 (starting from a folded state) and e2 (starting from a crumpled state). From the folded state, it is the left hand that is holding the corner and the right hand that slides, while for e2 is the right hand that holds and the left hand slides. In all cases, the hand that holds moves up to remove the table contact (see trajectories of e1 and e2 in \autoref{fig:actionGraph2}-right), and the sliding hand also moves up at the beginning of the motion, specially for the big tablecloth object. 

The second conclusion is that in most cases edge tracing is used not to flatten but to reach the corner. In other words, the subject first goes to reach the corner, and once the corner is reached the subject goes to the central state (named 2PP-LC+RC-Flat, also appearing in \autoref{fig:graphRestricted}-(c)). This can be seen in the trajectories of e1 and e2 (\autoref{fig:actionGraph2}-right) because the end points of the sliding hand are always much lower than the holding hand.  However, for some instances the subject goes to the central position while doing the edge tracing. This requires more ability because there is more risk of losing the grip, therefore appears only a few times in the data, and it's shown in the trajectories of e3 all corresponding to the small napkin that is an easier object to manipulate.

Finally, we can also observe how sometimes edge tracing doesn't reach the other corner. Then, either an additional edge tracing was performed to reach the corner (after shaking the cloth) or it continued with the following transition to place it on the table. Therefore, this shows us that even without achieving a task as expected, the final task can be done successfully, which is relevant when designing benchmarks for cloth manipulation.

 \section{Discussion}
 
%Graphs can be easily used to high-level planning decision making. In addition, we have the trajectories, velocity and acceleration profiles of all the  trials corresponding to each of the edges. 
To build the CLoM graph the proposed granularity for segmentation is much thinner than other works like \cite{lippi2020latent }, where only states with the cloth on the table are considered. This is done with the motivation of obtaining simple motion primitives to facilitate re-usability. % that can be reused in many tasks. The re-usability of primitives will depend on how much they depend on the actual particular cloth configuration that now we just consider in categories. 
In addition, we believe this segmentation is also relevant for benchmark purposes, to represent the complexity of a task and identify different evaluation segments.

Another motivation behind our approach is to enable explainable reasoning at the manipulation level as well as learning a dynamic movement primitive (DMP) for each re-grasp strategy (not necessary from human motion data), which is also associated with its preconditions and effects. The resulting DMPs can be used for task planning \cite{canal2018joining}, and potentially for explainability purposes as well, since the learning process makes explicit the conditions that enable to execute the primitive and the expected outcomes. We envisage the CloM Graph as a common ground representation where information at the different robotic levels (planning, perception and execution) can be stored.

In addition, the state representation simplifies the perceptual information that needs to be acquired. Thus, in subsequent research, we plan to use previous work in our group on cloth part recognition and pose estimation \cite{ramisa20163d} and grasping point detection \cite{corona2018active} to perceive the aforementioned manipulation-oriented scene states, including cloth state, grasping point location and confidence values that can provide explanations about the belief in the current state.

%In future work, we will work on classifying trajectories and learning low level movement primitives to ground this symbolic plans to executed them on robots, using  our group experience \cite{colome2018dimensionality,}.

\section{Conclusions}
We have introduced a compact and generic representation of states of a manipulation task in the context of cloth manipulation. The representations are vast simplifications of the complexity of a cloth manipulation state, but we showed how they are enough to segment a manipulation task into relevant and coherent manipulation primitives. In addition, from the sequences of states and transitions, we have defined the CloM Graph that encodes the diversity of strategies to accomplish the task. We have shown two examples of common cloth manipulation tasks for which the CloM Graph is learned from an experiment with 8 human subjects. Learning from human demonstrations allows to identify manipulation primitives not used so far by robots that could be especially handy for the versatile manipulation of clothing items.

The CloM Graph we have proposed complies with the desideratum that "low-complexity representations for the deformable objects should be the objective" \cite{smith2012dual}. This manipulation-oriented representation would permit probabilistic planning of actions to ensure reaching the desired cloth configuration without requiring high accuracy in perception nor searching in high-dimensional configuration spaces. 
In addition, our encoding of manipulation tasks facilitates the definition of metrics and measures of complexity of a given strategy, which is very useful to define benchmark tasks with increasing complexity. 

In future research, we will work towards the state recognition and the definition of the motion primitives performing transitions between states. 
Additionally, this work will lead to a database of labelled video data synchronized with motion data of different cloth manipulation tasks, which could be of great utility for the whole manipulation community working on highly deformable objects. 
%\addtolength{\textheight}{-10cm}

%Also, from our definition of states that represent a task, concrete requirements can be extracted to guide gripper design.
%

%We define such generic representations with the idea that they can facilitate state recognition, learning task strategies, manipulation movement primitives and symbolic planning, that is, they can be used to interchange information across the different layers of a robotic system. 

%Our aim is to learn the map of possible re-grasps by capturing motion data from humans manipulating textiles using wearable grippers, thus limiting human dexterity to that of the robot gripper design.

%- Un article que potser et pot ser útil:
%https://web.cs.wpi.edu/~rich/heres_how/pub/MohseniEtAl2016_ROMAN.pdf  
%(Conclusions - utilitat dels grafs etiquetats de manipulacions fetes per persones)

\bibliographystyle{ieeetr}

\end{document}